\documentclass[sigconf]{acmart}

\AtBeginDocument{%
  }

\copyrightyear{2022} 
\acmYear{2022} 
\setcopyright{acmcopyright}\acmConference[MM '22]{Proceedings of the 30th ACM International Conference on Multimedia}{October 10--14, 2022}{Lisboa, Portugal}
\acmBooktitle{Proceedings of the 30th ACM International Conference on Multimedia (MM '22), October 10--14, 2022, Lisboa, Portugal}
\acmPrice{15.00}
\acmDOI{10.1145/3503161.3548315}
\acmISBN{978-1-4503-9203-7/22/10}

\usepackage{enumitem}
\usepackage{dsfont}
\usepackage{pifont}
\usepackage{multirow}
\usepackage{amsfonts}
\usepackage{makecell}

\begin{document}
\title{Self-supervised Scene Text Segmentation with Object-centric Layered Representations Augmented by Text Regions}



\author{Yibo Wang}
\orcid{0001-5636-4485}
\affiliation{%
	\institution{School of Software Engineering, University of Science and Technology of China}
	\city{}
	\state{}
	\country{}
}

\author{Yunhu Ye}
\affiliation{%
	\institution{School of Software Engineering, University of Science and Technology of China}
	\city{}
	\state{}
	\country{}
}

\author{Yuanpeng Mao}
\affiliation{%
	\institution{School of Software Engineering, University of Science and Technology of China}
	\city{}
	\state{}
	\country{}
}

\author{Yanwei Yu}
\authornote{Corresponding Author}
\affiliation{%
	\institution{School of Software Engineering,  University of Science and Technology of China,\\Suzhou Institute for Advanced Research, University of Science and Technology of China}
	\city{}
	\state{}
	\country{}
}

\author{Yuanping Song}
\affiliation{%
	\institution{School of Software Engineering, University of Science and Technology of China}
	\city{}
	\state{}
	\country{}
}

\renewcommand{\authors}{Yibo Wang, Yunhu Ye, Yuanpeng Mao, Yanwei Yu, Yuanping Song}
\renewcommand{\shortauthors}{Yibo Wang et al.}

\begin{abstract}
	Text segmentation tasks have a very wide range of application values, such as image editing, style transfer, watermark removal, etc.
	However, existing public datasets are of poor quality of pixel-level labels that have been shown to be notoriously costly to acquire, both in terms of money and time. At the same time, when pretraining is performed on synthetic datasets, the data distribution of the synthetic datasets is far from the data distribution in the real scene. These all pose a huge challenge to the current pixel-level text segmentation algorithms.
	To alleviate the above problems, we propose a self-supervised scene text segmentation algorithm with layered decoupling of representations derived from the object-centric manner to segment images into texts and background. 
	In our method, we propose two novel designs which include Region Query Module and Representation Consistency Constraints adapting to the unique properties of text  as complements to Auto Encoder, which improves the network's sensitivity to texts.
	For this unique design, we treat the polygon-level masks predicted by the text localization model as extra input information, and neither utilize any pixel-level mask annotations for training stage nor pretrain on synthetic datasets.
	Extensive experiments show the effectiveness of the method proposed. On several public scene text datasets, our method outperforms the state-of-the-art unsupervised segmentation algorithms.
\end{abstract}




\maketitle

\section{Introduction}
With the development of convolutional neural networks in the past decades, scene text localization algorithm \cite{qin2021mask,long2018textsnake,dai2021progressive,liao2020mask} aiming to localize all text instance regions on a natural image has become mature. However, scene text segmentation differs from scene text localization which is considered as a coarse segmentation task utilizing polygon-level region mask annotations, while the former requires extremely large-scale pixel-level mask labels. Although the text segmentation task requires more elaborate mask labels, it has broader real-life applications such as image inpainting \cite{pathak2016context}, text style transfer \cite{yang2019controllable}, and watermark removal \cite{liang2021visible}. 
Therefore, although obtaining pixel-level labels is time-consuming, due to the prominent role of scene text segmentation in both industry and research communities, there is no doubt that we need to promote text segmentation algorithms development.

However, current text segmentation algorithms suffer from various problems. The first is the enormous diversity of data distribution. A scene text image has widely varying color, font, and shape gaps between text and background. Besides, existing public text segmentation datasets, including TextSeg \cite{xu2020rethinking} , TotalText \cite{ch2017total}, COCO-Text \cite{bonechi2019coco_ts}, ICDAR-13 \cite{karatzas2013icdar}, and MLT-S \cite{bonechi2020weak} are scarce and most of them have their limitations due to the small scale of the dataset either or a large scale contained many noisy annotations generated by machines. 
At the same time, early works \cite{wang2021semi,bonechi2020weak} argue that, unlike the text localization task that can use synthetic datasets for training, the synthetic data distribution of the text segmentation task on the Synth80K \cite{gupta2016synthetic} dataset is too far from the real-world text segmentation data distribution. So large-scale pretraining like ImageNet \cite{deng2009imagenet} is not feasible. 
\begin{figure}[htb]
	\centering
	\includegraphics[width=\linewidth]{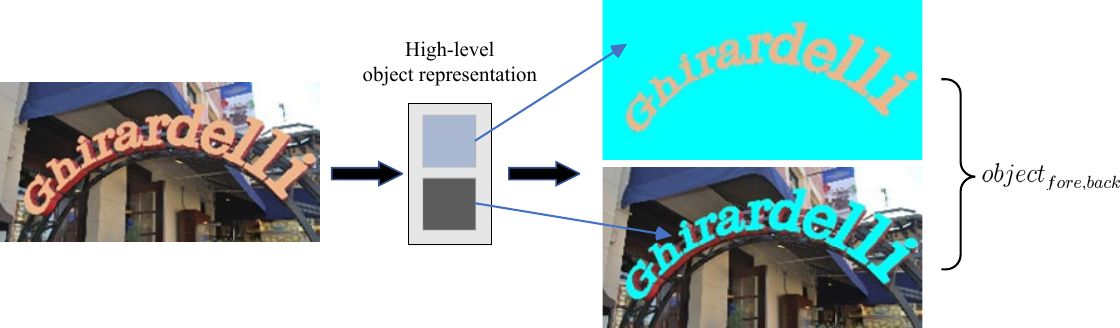}
	\caption{An image is segmented into two objects (foreground and background), represented by cyan. Each object is decoded from a high-level representaion.}
	\label{supervised-compare}
\end{figure}

Nowadays, the conception of object-centric \cite{russakovsky2012object} manner recently brings much inspiration to researchers. It heuristically believes that if objects with different components are to be split, the same component should be clustered together on high-level semantic features, while different components have a larger semantic gap in high dimensions. 
As shown in Figure \ref{supervised-compare}, the object-centric method utilize different objects to decouple and reconstruct to obtain segmentation masks, which can be regarded as an annotation-free method.
Therefore, inspired by object-centric conception, this paper aims to propose a self-supervised text segmentation algorithm whose core idea is object-centric perceptual aggregation to avoid the use of scarce high-quality pixel-level text segmentation labels. We decouple the original image into text and background. 

In particular, to acquire better representations, we propose two designs, region query and representational consistency, specific to scene text task. 
We receive inference results (text regions)  from an external text localization model. This brings a strong signal that we can suppress more false positives than general object detection tasks and focus more on local content.
Before gathering pixel features, the model queries the region so that the model's attention is focused on the foreground text. In addition, by replacing the irrelevant background outside the text area, the foreground features gathered before and after the replacement are guaranteed to be consistent so that the models can learn consistent higher-level semantic information.
We embed the above designs into Auto Encoder \cite{kingma2013auto} for self-supervised training, which constitutes our proposed method. The self-supervised signal is generated by reconstruction loss between a simple linear weighted summation of the two decoupled objects (text and background) and the input image. 

In the end, the main contributions of this paper are as follows:
\begin{itemize}
	\item We design the first self-supervised scene text segmentation algorithm that does not use any pixel-level labels based on the idea of object-centric method. At the same time, this method can perform segmentation without pretraining on any synthetic datasets.
	\item We add auxiliary specific learning signals which enhance the foreground representation unique to scene text segmentation, including two key innovations, \textbf{R}egion \textbf{Q}uery \textbf{M}odule, and \textbf{R}epresentation \textbf{C}onsistency \textbf{C}onstraints.
	\item Experiments show that our proposed algorithm outperforms the state-of-the-art unsupervised segmentation algorithms on three public scene text datasets.
\end{itemize}
\begin{figure*}[!t]
	\centering
	\includegraphics[width=\textwidth]{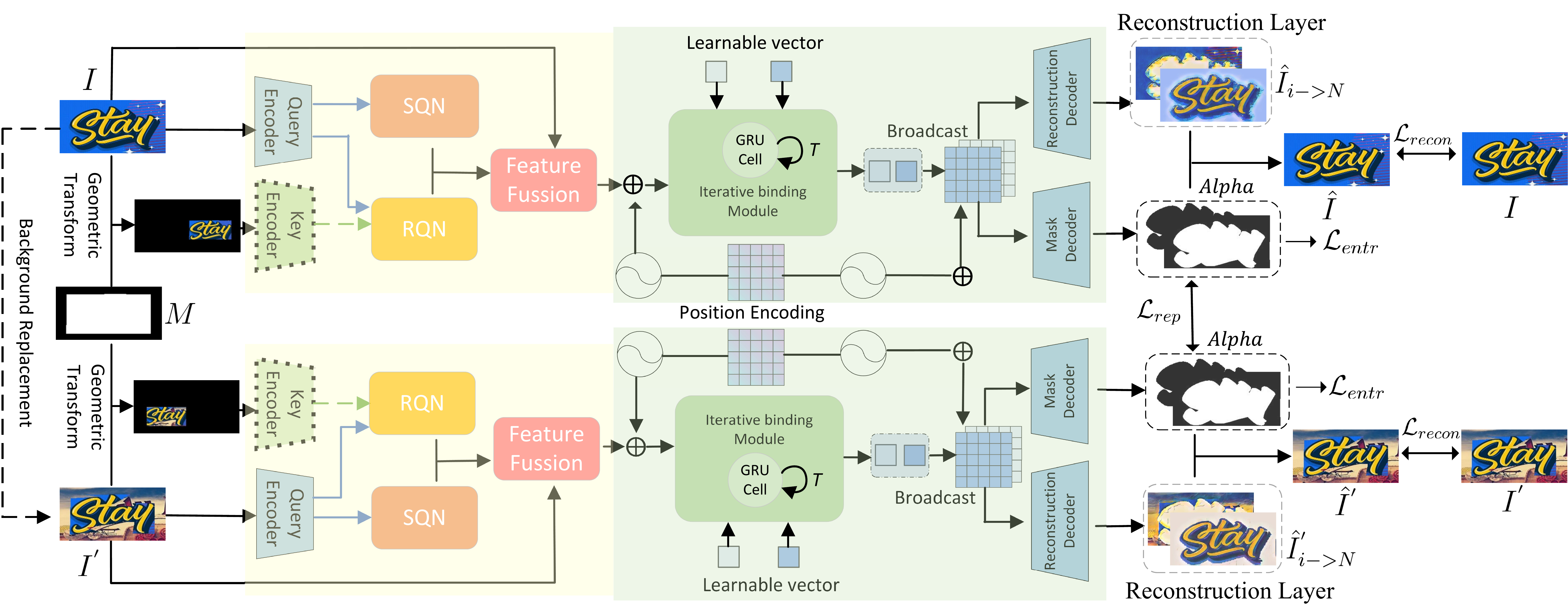}
	\caption{The overall pipeline of our proposed method. The whole illustration can be divided into two parts: the region query module (yellow part) and the feature binding module (green part). It contains upper and lower parallel parts, which together constitute the background replacement, and they are allocated in the same batch. }
	\label{pipeline}
\end{figure*}
\section{Related works}
\noindent \textbf{Scene Text Segmentation:} The goal of text segmentation is to segment each character from the background by finding its boundary. In the early days, text segmentation algorithms followed the way of image processing \cite{otsu1979threshold,su2010binarization}. However, they need to set a threshold for each image, which is inefficient. In addition, \cite{tang2017scene} uses multi-stage processing and statistical methods for text segmentation. Deep learning methods are more efficient than the tedious artificial threshold and parameter methods mentioned above. \cite{xu2020rethinking} is a recently proposed text segmentation algorithm specific to scene text datasets, which utilizes a series of auxiliary supervision signals to make segmentation better than existing general semantic segmentation algorithms. Both \cite{bonechi2020weak,wang2021semi} are weakly supervised text segmentation algorithms. The former uses synthetic datasets for pretraining and then transfers to real datasets, while the latter uses text region masks and text pixel masks to guide each other for training. Although the above two weakly supervised methods can generalize well, they both use pixel-level masked supervision signals in the training phase. Compared with them, our method can be regarded as zero-label text segmentation algorithms that do not use pixel-level masks or need to be pretrained on synthetic data.

\noindent \textbf{Self-Supervised Learning based on Auto Encode:} Self supervision can be used to learn valuable representations for downstream tasks without labeling data. It is often used as a large-scale pretraining model. After achieving significant success in NLP \cite{DevlinCLT19,brown2020language}, a large number of extraordinary research results have begun to emerge in the field of computer vision \cite{chen2020simple,grill2020bootstrap,he2020momentum,van2018representation}, such as image inpainting\cite{pathak2016context}, image restoration \cite{chen2019self}. Benefit from end-to-end learning, self-supervised learning based on generative model AE (Auto Encoder \cite{kingma2013auto}), can well integrate downstream tasks (\emph{e.g.} segmentation, object detection) into self-supervised tasks. Based on AE, there are also many exciting applications. \cite{wang2021unsupervised,kulkarni2019unsupervised} are proposed to use the AE framework for self-supervised training to model the visual environment to improve the environmental anti-interference ability of reinforcement learning\cite{kaelbling1996reinforcement}. \cite{zoran2021parts,locatello2020object} are based on AE to synthesize virtual view data after training on video stream data and perform unsupervised segmentation of images.

\noindent \textbf{Weakly/Unsupervised Semantic Segmentation: }Different from fully supervised semantic segmentation methods \cite{chen2017deeplab,fu2019dual,yuan2021ocnet,wang2021max}, \cite{liu2021weclick,obukhov2019gated,zhang2020weakly} are proposed to use weak labels for semantic segmentation of images because of the high cost of pixel-level label acquisition. They use object clicks or skeleton labels to train the model outperforming fully supervised training manners. Although they have lower acquisition costs, the hyperparameters of the weak label area have to be artificially selected. In order to solve the tedious process of artificial selection, \cite{nguyen2019deepusps,shimoda2019self} propose to use the properties of the image itself, such as histogram and other hardcore methods, to first process the image to obtain inaccurate pseudo-labels, then employ pseudo-labels as labels for supervised learning while filtering out the errors of segmentation results. However, the above methods are all solutions without human intervention, so \cite{locatello2020object,yang2021selfsupervised} propose a fully self-supervised segmentation method to obtain different objects of the image. Our method's framework is the object-centric self-supervised semantic segmentation proposed by \cite{locatello2020object}. Since object-centric algorithms can well represent objects as high-level semantic features, they are relatively suitable for self-supervised semantic segmentation.

\noindent \textbf{Layered Representation:} It was initially proposed in \cite{wang1994representing} to decompose a video into a series of simpler motion levels. Since then, layered representations have been widely used in computer vision \cite{jojic2001learning,pawan2008learning}. Besides, deep learning based layered decoupling methods is used to do virtual scene synthesis \cite{locatello2020object}, optical flow layering \cite{yang2021selfsupervised}, or separation of foreground/background \cite{gandelsman2019double}. It is now widely believed that segmentation tasks are divided into top-down \cite{tian2021boxinst,zhao2017pyramid,long2015fully} and bottom-up \cite{choudhury2021unsupervised,monnier2021unsupervised}. The former is a conventional segmentation method that has emerged with the advancement of deep learning in recent years and uses the supervision signal to classify each pixel. The latter is similar to clustering \cite{macqueen1967some}, starting from each pixel and clustering their similar parts, such as color, semantic relationship, optical flow, \emph{etc}, in the neighborhood of a higher-level semantic space. Because the segmentation algorithm based on self-supervised learning does not have a label for each pixel, the bottom-up segmentation method is more suitable for self-supervised semantic segmentation.

\section{Our Method}
We train a self-supervised model framed by a generative model from scratch. As shown in Figure \ref{pipeline}, First, the text segmentation model utilizes the cropped image and corresponding region image (all black outside the text region) as input and encodes them at the same time. After the objects are successfully decoupled, the layered coefﬁcients are used to reconstruct the original image with the layered objects. Since the content of the encoded foreground region is of interest, during the encoder phase, we pass a region query module to query the specified foreground. For its benefit, the model's attention focuses on the text region during decoupling.  Additionally, we use segmentation consistency after background replacement to improve the accuracy of self-supervised signals. 

\begin{figure*}[!t]
	\centering
	\includegraphics[width=\textwidth]{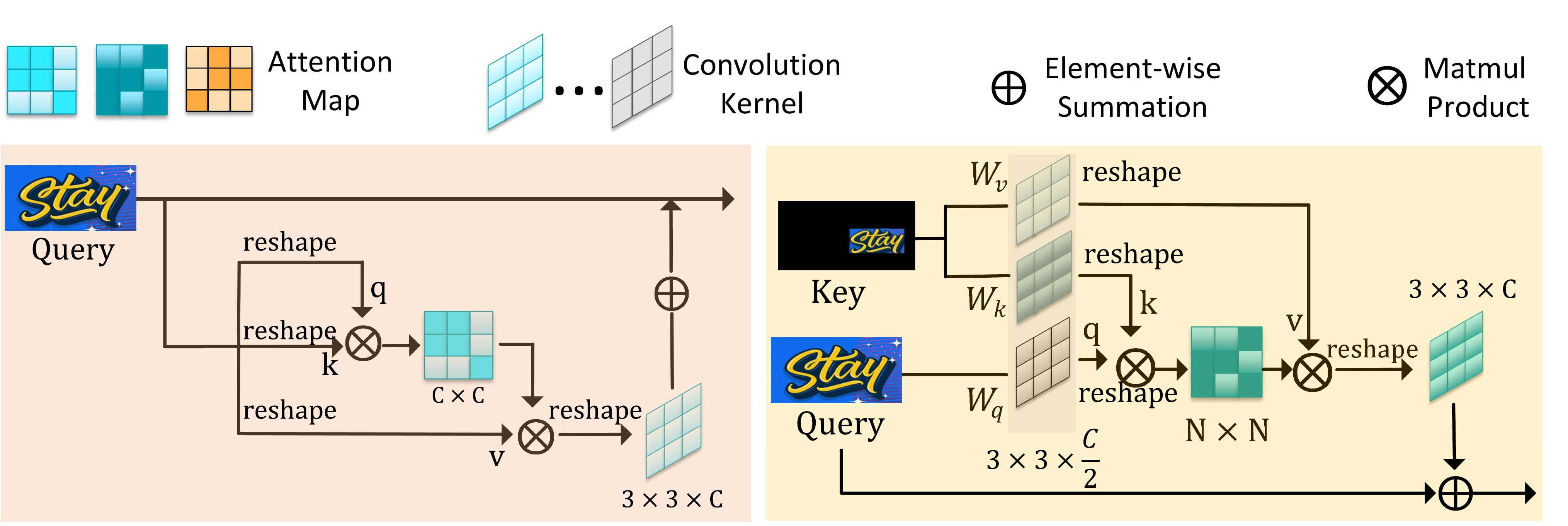}
	\caption{From left to right of illustraition are the Self Query Network (SQN) and Reference Query Network (RQN).}
	\label{rqm}
\end{figure*}
\subsection{Object-Centric Layered Representation}
For a natural image, it is composed of multiple components, and each component object is linearly weighted summation through the $Alpha$ channel (mask) to make up the image. The layered representation proposed in our paper is based on this object-centric method. We argue that text images can be composed of two objects, the text foreground, and the text-irrelevant background:
\begin{equation}\label{equ:DisentangleImage}
	\phi(I) = {\left \langle \alpha_i,\hat{I}_i \right \rangle}_{i=1}^K  \qquad
	\hat{I} = \sum_{i=1}^{K}{\alpha_i \cdot  \hat{I}_i} \qquad
	\mathcal{L}_{recon} = MSE\left(I,\hat{I} \right)
\end{equation}
where $I$ represents the original image, $\alpha_i$ represents the $Alpha$ mask of the $i^{th}$ layer obtained by inputting the image into the whole pipeline of Figure \ref{pipeline}, $\hat{I}_i$ represents the reconstructed image of the $i^{th}$ layer, and $\phi$ represents the whole layered decoupling pipeline. In our structure, $K$ is 2, and then recombine the output results of the model to obtain the reconstructed image $\hat{I}$ of the input image $I$. Then calculate the mean square error $\mathcal{L}_{recon}$ between them as the self-supervised training signal.

As shown in the Figure \ref{pipeline} green part, representation iterative binding module exploits the similarity among pixels to group them into the same high-level semantic feature.
As clustering algorithms do, we iteratively cluster the features of the image. To implement this idea, we use a method similar to the Slot Attention\cite{locatello2020object}. In the original Slot Attention, a Gaussian distribution is used to represent each object. However, here we use two learnable query vectors $Q \in \mathbb{R}^{D \times \ K}$ as slots, like DETR \cite{carion2020end} does, where $D$ is the encoding dimension of each layer. Each query vector represents a layered object through the softmax-based attention mechanism, and these slots are iteratively updated using a loop update function. In scene text segmentation, each query vector represents the encoding of text foreground or irrelevant background, respectively, which are then decoded and recombined to reconstruct the image. The input of the iterative binding module is the output feature map $F$ of the feature extractor, we denote the query vector $Q$ as $q$, and the feature map $F$ as $k, v$, we compute the binding weights:
\begin{equation}\label{equ:Attn}
	\begin{split}
		{attn}_{i,j}^t = \frac{\exp(M_{i,j}^t)}{\sum_{l=1}^{K}\exp(M_{i,l}^t)} \quad  where \quad M^t = \frac{k^T \cdot q^t}{\sqrt{D}} \in \mathbb{R}^{N \times K}
	\end{split}
\end{equation}
where $K$ is set to 2, $q^t$ is the query result of step $t$ in the binding process, unlike the standard Self Attention, which calculates attention weights in the pixel dimension, we decompose the features in the slot dimension. This is chosen because we want to bind the representations hierarchically by making them compete to ensure that each pixel is well assigned to a different or the same query vector. We can derive the formula for pixel clustering:
\begin{equation}\label{equ:DistangleAttn}
	\begin{split}
		{U}^t = v \cdot A^t \in \mathbb{R}^{D \times K} \quad where \quad A_{i,j}^t = \frac{{attn}_{i,j}^t}{\sum_{l=1}^{N}{attn}_{l,j}^t} 
	\end{split}
\end{equation}
where $N$ represents the number of pixels. Due to the competing process, the binding of representations also works well to explain different parts of the input image. The above formula expresses one iteration execution, but only one iteration cannot obtain enough semantic information. So as shown in Figure \ref{pipeline} green part, in order to maintain a smooth iterative update, we use a Gated Recurrent Units (GRU) \cite{cho2014learning} as the iterator of the query vector:
\begin{equation}\label{equ:GRU}
	Q^{t} = GRU(Input = U^t,Memory = Q^{t-1})
\end{equation}
the entire iterative step is bound to $T$ steps, in our paper $T=5$. Finally, we want the values of the mask to be as close to 0 or 1 as possible while the probability sum is 1, so we use entropy loss as follows:
\begin{equation}\label{equ:EntropyNorm}
	\mathcal{L}_{entr} = -\frac{1}{\Omega}\sum_{\forall i\in K}\sum_{\forall I_j \in \Omega}{\alpha_i(I_j)} \cdot \log{\alpha_i(I_j)}
\end{equation}
the $\mathcal{L}_{entr}$ forces the mask channel to approximate the $One-Hot$ vector to make the pixels gathered into the foreground closer to the foreground and the background closer to the background. We minimize this loss and make each of them close to binarization.
\subsection{Text Region Query Module}
Text Region Query Module, as a feature extractor for representation binding module, is divided into two parts, Region Encoding Module and Region Query Module. As Figure \ref{pipeline} yellow part shows, the REM provides the prerequisite coding conditions for the RQM, and the RQM integrates the Query and Key networks to provide augmented semantic features for the binding module.

The REM is divided into two branches, Query and Key network. The former encodes the original input image, while the latter encodes the region image. Because the image fed into the Key network contains a lot of black areas, these black areas will bring a lot of zero elements which will cause the model to collapse. According to the processing method of MOCO \cite{chen2020improved}, we freeze the parameters of the Key network and use the parameters of the Query network to momentum update them. Since we truncate the gradient of the Key network, the gradient of the downstream network cannot affect its parameters.

Inspirited by \cite{fu2019dual}, our RQM consists of two components, Reference Query Network and Self Query Network. The RQM provides two functions, "where to look" and "what to see", which are provided by RQN and SQN, respectively. Since RQN only queries the location information, the semantic features it captures are foreground instance matching. However, if only the location matching information is employed, the model cannot learn enough effective content well, so we design SQN in parallel.

\noindent \textbf{Reference Query Network :}\ As the right part of Figure \ref{rqm} shown, we regard the output result of the Key Encoder and the Query Encoder $K, Q \in  \mathbb{R}^{\frac{H}{\eta}\times {\frac{W}{\eta}\times D}}$ as referenced object $k,v$ and encoding object $q$ respectively. In our design, we do not perform geometric data augmentation on the input of the Query Encoder, but for the input of the Key Encoder, we perform data augmentation with flipping, translation, and zoom-out. We calculate the similarity of spatial dimensions. First, we flatten the ${h,w}$ dimensions. After flattening, the feature map becomes $q,k,v \in \mathbb{R}^{N \times D}$, where the size of $N$ is $h \times w$, $ h$ and $w$ are 1/$\eta$ times the height $H$ and width $W$ of the original image. $\eta$ is set to 4 in our paper. We have proved through experiments that the size of the feature map when performing region query and representation binding should not be too small, otherwise important semantic information will be lost. So in our paper, we downsample it to 1/4 times the size of the original image. We directly calculate the similarity in the $N$ dimension:
\begin{equation}\label{equ:ReferenceQueryAttn}
	A_{rqn} = S \cdot v + q  \qquad
	{sim}_{ij} = \frac{\exp((q_i \cdot k^T_j) / \sqrt{D})}{\sum_{i=1}^N{\exp((q_i \cdot k^T_j) / \sqrt{D}))}}
\end{equation}
where ${sim}_{ij} $ represents the similarity between position $i$ and position $j$ of $S \in \mathbb{R}^{ N \times N }$. We let the encoding object (original image) query the referenced object (region image), which brings spatial information for Query Encoder.

\noindent \textbf{Self Query Network :}\ Unlike RQN, as shown in the left part of Figure \ref{rqm}, the referenced object and the encoding object of SQN are both from the same object, Query Encoder. As the name implies, SQN extracts the attention weights of semantic features by calculating the similarity among pixels of the same image. Here we abuse the notation definition. Like the definition of RQN, the feature map is defined as $Q,K \in \mathbb{R}^{\frac{H}{\eta}\times {\frac{W}{\eta }\times D}}$.
In the same way, we flatten the dimensions of $h,w$ to get $q,k,v \in \mathbb{R}^{N\times D}$, transpose $q$ to get $q^T \in \mathbb{R}^{ D\times N}$ and directly calculate the similarity in channel dimension:
\begin{equation}\label{equ:SelfQueryAttn}
	A_{sqn} = v \cdot S + q \qquad
	{sim}_{ij} = \frac{\exp((q_i ^T\cdot k_j)/\sqrt{D})}{\sum_{i=1}^N{\exp((q^T_i \cdot k_j)/\sqrt{D})}}
\end{equation}
the similarity here ${sim}_{ij} $ calculates position $i$ and position $j$ of $S \in \mathbb{R}^{D\times D}$ in the channel dimension $D$ similarity of location. Query, key, and value of SQN all come from Query Network. 

RQN obtains the query result in the spatial dimension, and SQN obtains the query result in the channel dimension. Also, we introduce the original image as an auxiliary interpretation object bringing low-level features. We are not directly doing element-wise summation but concatenating according to dimension among them.
\begin{figure}[!t]
	\centering
	\includegraphics[width=\linewidth]{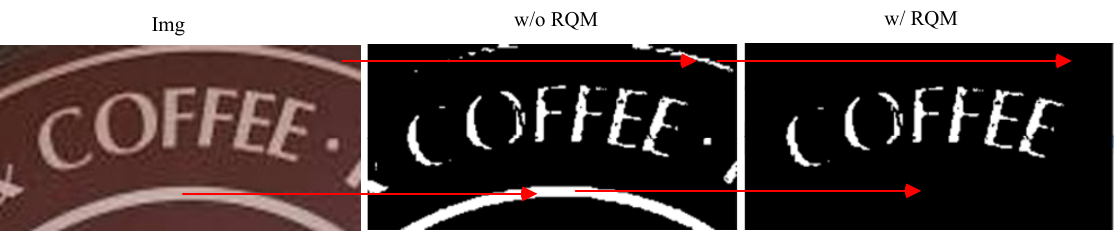}
	\caption{Before and after using RQM, it can be seen that the network does effectively suppress false positives.}
	\label{sup}
\end{figure}
\subsection{Representation Consistency Constraint}
The model can only learn relatively simple semantic relationships in the usage of low-level semantic information. Generally, there is bound to be a large gap in color between the foreground and the background, so the model is easy to overfit in color learning. For instance, the foreground is white, and the background is a mixture of blue and white. Since the model learns color information, the white part in the background is also regarded as part of the foreground, which causes the interference of false positives. Accordingly, we propose a solution called Background Replacement Representation Consistency Constraint.

The reason why background replacement can heighten semantic information is that for the foreground information we need, it should have a semantic invariance, which brings us the neglect of background. So if there are two images having the same foreground components while totally different backgrounds, they should have the same foreground mask. As shown in Figure \ref{pipeline}, we randomly select an image as the background and use the region obtained by the text localization task as the basis for background replacement. 

\noindent \textbf{Bulid Replacement:}\ Select an original image $I$ in the training set as the foreground image, and then randomly select a training set image $B$ as the background image. If text region is geometrically transformed, the region mask consistency will be destroyed. For this reason, we only perform the transformation on the color. The foreground and background images perform an element-wise summation operation depending on the text region mask $M$ to obtain the image $I^\prime$ after replacing the background. The two images are simultaneously fed into the model for segmentation:
\begin{equation}\label{equ:Background}
	I^\prime = I \odot M + B \odot (1-M)
\end{equation}

\noindent \textbf{Consistency Constraint:}\ We feed the image pairs constructed in the building step into the model and output the reconstruction layers and $Alpha$ masks of each image. Since the reconstruction loss is commutative, there is no guarantee that the same object will always output the background or foreground layers. So we use the Hungarian matching algorithm \cite{2010The} used in most object detection models to calculate the permutation-invariant consistency loss, which is only backpropagated through the lowest-error permutation:
\begin{equation}\label{equ:ConsLoss}
	\begin{split}
		h(i,j) = \frac{1}{\Omega}\sum_{\forall p \in \Omega}{\|\alpha_i(p) - \alpha^{\prime}_j(p) \|}_2 \\
		\mathcal{L}_{rep} = \mathds{1}_{\min\left \{h(i,j)\right \}}\cdot h(i,j) \quad where \quad i,j \in \Psi
	\end{split}
\end{equation}
where $h(i,j)$ calculates the Hungarian matching value between the original mask $\alpha$ and the replacement mask $\alpha^\prime$, and $\Psi$ represents different mask objects. Consistency loss only occurs in the training phase, so we do not use the replacement map to replace the background in the inference phase. We only use the original image and region image for inference in the inference stage.

Finally, in the training phase, our total loss can be calculated as:
\begin{equation}\label{equ:TotalLoss}
	\mathcal{L}_{total} = \lambda_1\mathcal{L}_{recon} + \lambda_2\mathcal{L}_{entr} + \lambda_3\mathcal{L}_{rep}
\end{equation}
in our experiment, it is specified that $\lambda_1, \lambda_2, \lambda_3$ are ${10}^{2}$, ${10}^{-2}$, ${10}^{-2}$ respectively. Using this staged training strategy brings more stable performance for our experiments. We minimize $\mathcal{L}_{total}$ to minimize reconstruction loss $\mathcal{L}_{recon}$, entropy loss $\mathcal{L}_{entr}$ and representation consistency loss $\mathcal{L}_{rep}$.

In the inference phase, we use Dense-CRF \cite{krahenbuhl2011efficient}, which widely believed to play a more critical role in the self-supervised scenario than fully supervised scenario, to postprocess the mask results.
\section{experiments}
\begin{figure*}[!t]
	\centering
	\includegraphics[height=.3\textheight,width=\textwidth]{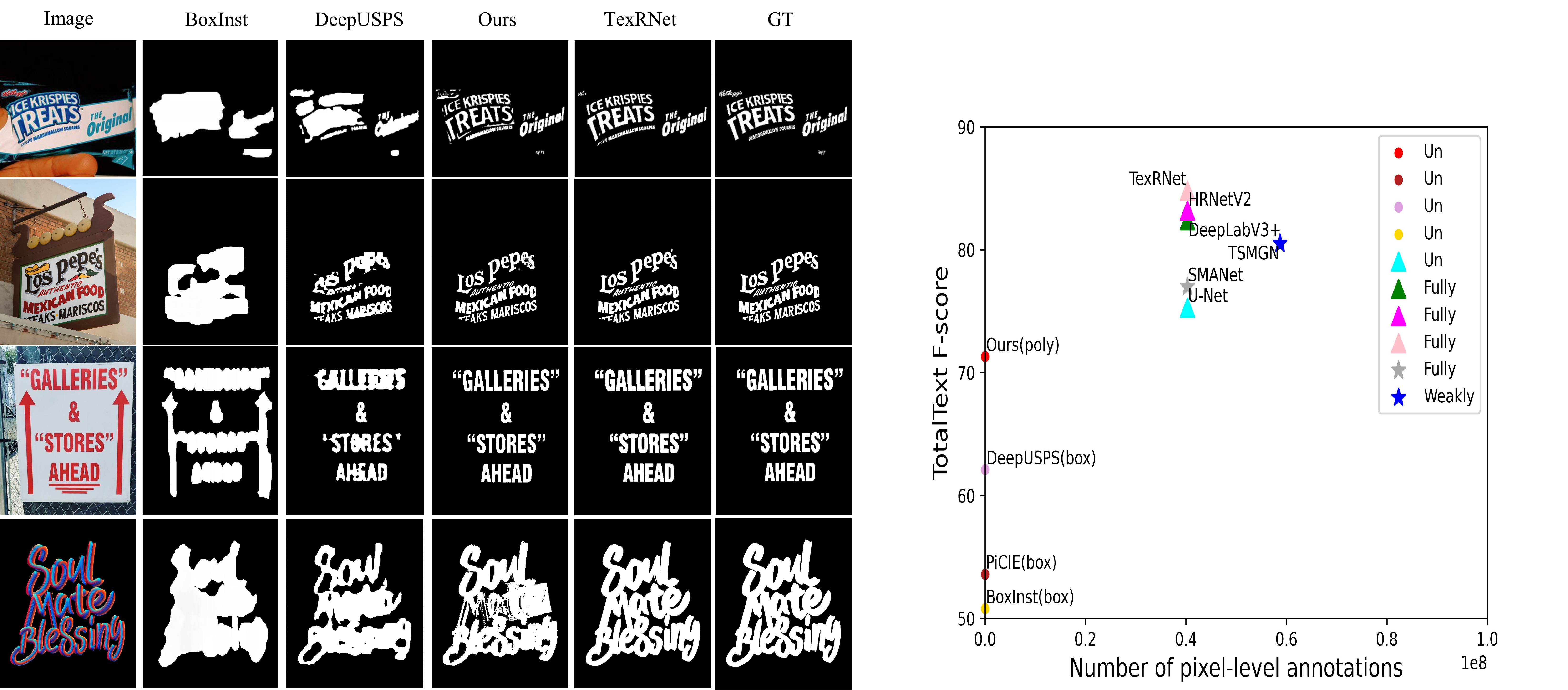}
	\caption{ (a)The top two rows are the results on the TotalText dataset, and the bottom rows are the results on the TextSeg dataset. (b) Pixel-level annotations comparison on TotalText. Some methods will be pretrained on Synthetic datasets, but here we only count the pixel-level annotations of TotalText. Different shapes of marks represent diverse types of supervision.}
	\label{result}
\end{figure*}
\subsection{Dataset And Evaluation Metrics}
Because the quality of pixel-level mask annotations of some public datasets such as COCO-Text and MLT-S is not high, resulting in inaccurate model performance evaluation, we only use them as supplementary image sets in our experiments and evaluate datasets with high mask quality (ICDAR-13, 
TotalText and TextSeg). 

In our experiments, we utilize two types of evaluation metrics. One of them is applied to evaluate comparison experiments, including F1-score, fgIoU, and accuracy. The first two are used to measure the segmentation performance of different models in Section 4.3, and accuracy is used to measure the text recognition performance of different models in Section 4.5. For another type of metric, we employed it to measure our ablation experiments under different conditions. We use F1-score to evaluate segmentation performance and exploit MSE to represent reconstruction error in Section 4.4.
\begin{table*}
	\centering
	\begin{tabular*}{\hsize}{@{}@{\extracolsep{\fill}}ccc|cc|cc|cc@{}}
		\toprule
		\multirow{2}{*}{Model} & \multirow{2}{*}{Ext} & \multirow{2}{*}{Supervised} &\multicolumn{2}{c}{TotalText} & \multicolumn{2}{c}{TextSeg} & \multicolumn{2}{c}{ICDAR-13} \\
		\cmidrule{4-9} 
		&&& fgIoU($\uparrow$)&  F1-score($\uparrow$)& fgIoU($\uparrow$)& F1-score($\uparrow$) & fgIoU($\uparrow$) & F1-score($\uparrow$)\\
		\cline{1-9}
		\midrule
		U-Net \cite{ronneberger2015u}     & \ding{55} & Fully   & $-$ & 0.753 &$-$ & $-$ &  $-$&0.708 \\
		DeeplabV3+ \cite{chen2018encoder}      & \ding{55} &Fully   & 74.4 & 0.824 &84.0&0.914& 69.2 &0.803\\
		HRNetV2 \cite{sun2019high}      & \ding{55}& Fully  & 76.2 & 0.832 &85.9& 0.918& 72.4 &0.830\\
		TexRNet \cite{xu2020rethinking}       & \ding{55} & Fully  & 78.4 & 0.848 &86.8& 0.924& 73.3 &0.850\\
		\cline{1-9}
		\midrule
		SMANet \cite{bonechi2020weak}       & S & Weakly+Box  & $-$  & 0.770 &$-$ &$-$ & $-$ &0.785 \\
		TSMGN \cite{wang2021semi}       & C+M & Weakly+Poly &  $-$  & 0.805 & $-$ & $-$ &  $-$  &0.745 \\
		\cline{1-9}
		\midrule
		BoxInst \cite{tian2021boxinst}      & \ding{55}& Un+Box  & 45.5 &0.508&39.5 & 0.542&33.0&0.477 \\
		PiCIE \cite{cho2021picie}      & \ding{55}& Un+Box  & 46.9 & 0.536&44.3& 0.513 &45.1& 0.548 \\
		DeepUSPS \cite{nguyen2019deepusps}  & \ding{55}& Un+Box  & 47.8 & 0.621 &58.4&0.722& 50.8 &0.657\\
		\cline{1-9}
		\midrule
		Ours(poly-det)    & \ding{55}& Un+Poly  & 65.1 & 0.713 &64.6& 0.725 & 53.3& 0.656  \\
		Ours(poly-gt)   & \ding{55}& Un+Poly  & 76.6 & 0.838 &74.3& 0.828 &65.8& 0.763 \\
		\bottomrule
	\end{tabular*}
	\caption{Performance comparison among our method and other methods on three text segmentation datasets. The second column, Ext means whether to use extended annotations for training. C, M and S represent COCO-Text, MLT-S and Synth80K datasets respectively. The third column represents whether to use pixel-level annotations to train, and Box (bounding box) and Poly (polygon region) mean additional cues as mentioned.}
	\label{Comparion}
\end{table*}
\begin{table}
	\begin{tabular*}{\hsize}{@{}@{\extracolsep{\fill}}cccccccc@{}}
		\toprule
		Model & ${model}_{det}$ &$\eta$& $A_{rqn}$ & $A_{sqn}$ & $L_{rep}$ & F1($\uparrow$) & MSE($\downarrow$)\\
		\midrule
		$\rm{Ours}$-$\alpha$   &  PCR &4&\ding{55}  & \ding{55}   &\ding{55}  & 0.490 & 7.02\\
		$\rm{Ours}$-$\beta$   & PCR &4& \ding{55} & \ding{55}  &\ding{51}  & 0.507&8.45\\
		$\rm{Ours}$-$\gamma$&  PCR &8&\ding{55} & \ding{55}      &\ding{51} & 0.459&9.63\\
		\midrule
		\cline{1-8}
		$\rm{Ours}$-$\delta$ & PCR &4& \ding{51}  & \ding{55}      & \ding{51} &  0.557&6.62 \\
		$\rm{Ours}$-$\epsilon$  & PCR &4& \ding{55}  & \ding{51}     & \ding{51} & 0.523&6.90\\
		$\rm{Ours}$-$\xi$  & PCR &4& \ding{51}  & \ding{51}     &  \ding{55} &0.594&7.92\\
		\midrule
		\cline{1-8}
		$\rm{Ours}$-$\zeta$&  TextSnake&4& \ding{51} & \ding{51}      &\ding{51} & 0.626&6.68\\
		$\rm{Ours}$-$\theta$&  DB&4& \ding{51} & \ding{51}      &\ding{51} & 0.709&\textbf{6.27}\\
		$\rm{Ours}$-$\kappa$&  PCR &4&\ding{51} & \ding{51}      &\ding{51} & \textbf{0.713} &6.31 \\
		\bottomrule
	\end{tabular*}
	\caption{Ablation experiments of our method on TotalText. The columns from left to right are region detection model (${model}_{det}$), downsample ratio ($\eta$), reference query network ($A_{rqn}$ ), self query network ($A_{sqn}$) and consistency loss ($L_{rep}$).}
	\label{Ablation}
\end{table}

\subsection{Implementation Details}
Our entire code runs on a NVIDIA Tesla V100 32G GPU. In the training phase, we crop out text region according to the bounding box of the text area and then unify the fixed image size as $128\times256$. We use data augmentation with flipping, translation, and zoom-out when the Key Encoder is used and transforms on HSV, which are applied in other cases. At the beginning of training, we use a learning rate of $5\times{10}^{-4}$ on the Adam optimizer, and at $2\times{10}^3$ iterations at the beginning of training, model is trained with warm-up. After that, we attenuate the learning rate by half for every ${10}^5$ iterations and simultaneously expand the auxiliary limit coefficients $\lambda_2$ and $\lambda_3$ by $5$ times. The total iteration steps of training are about $5\times{10}^5$. Our implementation purely uses a simple VGG \cite{simonyan2015very} network as our backbone. Moreover, our decoder network is not like that used in  \cite{locatello2020object} and \cite{yang2021selfsupervised}, which integrate mask generation and layer reconstruction into the same network. They use transposed convolution to get mask and layer objects, but in experiments, we found that this hurts the performance of both, so we separate them into upsampling and transposed convolution. The former is used as the mask decoder, and the latter is used as the decoder of object reconstruction. The segmentation results are shown in Figure \ref{result}.
\subsection{Experimental Results}
We compare our proposed algorithm with unsupervised segmentation algorithms and fully/weakly supervised segmentation algorithms, respectively. The comparison results on three datasets are shown in Table \ref{Comparion}. Note that in the last two rows of Table \ref{Comparion}, we discuss the two situations separately. We use the text region obtained by the text localization model, and the other is the ground truth of the text region. Here all comparisons are done with Dense-CRF.

\noindent \textbf{Comparison with un-supervised algorithms:}\ We compare our algorithm with three general segmentation algorithms now widely used without pixel-level label supervision. It should be noted that the third column of Table \ref{Comparion} represents whether to use pixel-level labels for training and whether to use additional cues. BoxInst, which uses the label information of bounding box for weakly supervised training in the training phase, is different from our algorithm, DeepUSPS, and PiCIE. Although we use additional information obtained through the upstream localization model, these cues do not participate in training and are only used to crop out the minimum bounding rectangle of text regions.
Since DeepUSPS and PiCIE initially performed unsupervised segmentation on the entire image, here we crop out rectangle regions detected by PCR-Net\cite{dai2021progressive} for better performance and fair experiments. 
As shown in Table \ref{Comparion}, compared to the other three unsupervised models, our model achieves at least 9.2\%, 0.3\% F1-score improvement on the TotalText and TextSeg datasets, respectively. The results were unsatisfactory when our model was run on the ICDAR-13 dataset. The possible reason for this is that the ICDAR-13 dataset has very low resolution, so our model cannot disentangle them well. Another reason is that the dataset scale is too small to train the model well.


\noindent \textbf{Comparison with weakly/fully supervised algorithms:}\ In Table \ref{Comparion}, we select three proposed algorithms specific to scene text tasks, TexRNet, SMANet and TSMGN. It can be seen that there is an evident gap between our model without pixel-level label supervision and the weakly/fully supervised learning models on all datasets. Since, unlike weakly/fully supervised algorithms that directly give the model pixel class supervision signals, self-supervised algorithms do not have pixel classification annotations. Only the mean square error between the reconstructed image and the original image is used to provide the supervision signal. 
As Figure \ref{result} shows, all three of the above methods utilize a large number of pixel-level labels.
Nevertheless, when we directly use ground truth of polygon mask, our model achieves some improvement. 
The localization model heavily influences our segmentation algorithm. When the performance of the localization model is better, the segmentation results will be closer to the results when polygon ground truth is given. In other words, if needing to perform pixel-level segmentation operations for a given region or generate new datasets with better quality pixel masks by polygon-level annotations, our model is currently a better choice.

\subsection{Ablation Study}
This section conducts ablation experiments on several vital points mentioned above. The dataset we choose is TotalText, and the results of our ablation experiments are shown in Table \ref{Ablation}.

\noindent \textbf{Localization model and downsampling size:}\ Our algorithm's performance is affected by text localization models, which cascadingly affect the region query module and background replacement. 
From the last three rows in Table \ref{Ablation}, there is an infinitesimal gap in segmentation results among good localization models \cite{liao2020real}. However, the result is extremely terrible when we use weaker localization methods \cite{long2018textsnake}. In addition, the downsampling scale of the encoding network will also impact the performance. From Table \ref{Ablation}, $\rm{Ours}$-$\gamma$ and $\rm{Ours}$-$\beta$, we use a larger feature map will bring 4.8\% F1-score improvement. Because the feature map of the last layer aggregates the semantic information with a larger window as the scale of downsampling increases, this will bring less similarity calculated among different pixels but more similarity in itself when binding, which degrades the performance of Slot Attention.

\noindent \textbf{Region query module:}\ We divide the region query module into $A_{rqn}$ and $A_{sqn}$ for ablation experiments. When we only consider the effect of self query network, as shown in Table \ref{Ablation}, the F1 score is improved by 1.6\% compared to $\rm{Ours}$-$\beta$. The insignificant improvement is that the self query network does not use any regional information, and the network pays attention to its channel information. When using only the reference query network, the F1 score gains improvement by 5\% compared to $\rm{Ours}$-$\beta$. The reference query network has an obvious effect because it distinguishs between text and background and solves the network's fuzzy identification between background and foreground. When we combine the reference query network and the self query network, as shown in Table \ref{Ablation}, $\rm{Ours}$-$\kappa$ has a relatively large improvement compared with no region query module. It makes sense because augmented semantics allow the network to learn content and spatial features simultaneously after the network has queried the region's location. 

\noindent \textbf{Representation consistency loss:} \ Although we propose a text region query module, it only enhances the primary semantic features of the network, and there is still a lot of low-level information obtained by the decoupling binding module. Because the general data augmentation method does not essentially enhance the semantic information of the data, we do not do geometric data enhancement here. Instead, we use the background replacement method to construct data pairs. Intriguingly, as shown by $\rm{Ours}$-$\alpha$ and $\rm{Ours}$-$\beta$ in Table \ref{Ablation}, when we only use background replacement without region query module, the improvement is limited, only 1.7\%, which may be due to the interference of a large number of false positives outside of text region when propagating the minimum error.
Compared with $\rm{Ours}$-$\beta$ and $\rm {Ours}$-$\alpha$, $\rm{Ours}$-$\kappa$ improves the F1-score by 20.6\% and 22.3\%, respectively.
It indicates that when we combine consistency loss with the text region query, the regional query network can help the network to suppress false positives, and they complement each other.
\subsection{Application}

The masks obtained by our proposed model can be well utilized in downstream tasks, such as text recognition and text style transfer.

\noindent \textbf{Improving text recognition:}\ We pass the text instance image (RGB) through our proposed model to get the mask (M), and then we concatenate the mask and greyscale of the image according to the channel dimension as input (grey-M) of the recognition model. Here as \cite{wang2021semi} does, we directly use the officially pretrained model of DAN \cite{wang2020decoupled}, and CRNN \cite{shi2016end} where the input layer is replaced, and other layers are finetuned. As shown in Table \ref{recshow}, according to our masks, the model does correct mistakes.
\begin{table}[htb]
	\centering
	\begin{tabular}{@{}@{}c|c|c|c@{}}
		\toprule
		Model&Image/Mask&GT&Pedict\\
		\midrule
		Ours(CRNN)& \makecell[c]{
			\begin{minipage}[b]{0.2\columnwidth}
				\centering
				\raisebox{-.5\height}{\includegraphics[width=\linewidth,height=.25\linewidth]{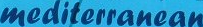}}
			\end{minipage} \\
			\begin{minipage}[b]{0.2\columnwidth}
				\centering
				\raisebox{-.8\height}{\includegraphics[width=\linewidth,height=.25\linewidth]{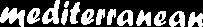}}
			\end{minipage}
		} & \raisebox{-.2\height}{mediterranean}& \makecell[c]{
			\raisebox{-.5\height}{mediter\textcolor[rgb]{1,0,0}{\_}anean}\\(no mask)\\ \raisebox{-.5\height}{mediterranean}\\(mask)
		} \\
		\midrule
		Ours(DAN)& \makecell[c]{
			\begin{minipage}[b]{0.2\columnwidth}
				\centering
				\raisebox{-.5\height}{\includegraphics[width=\linewidth,height=.25\linewidth]{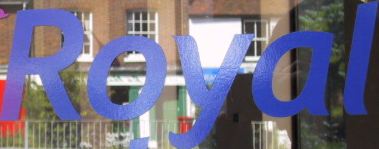}}
			\end{minipage} \\
			\begin{minipage}[b]{0.2\columnwidth}
				\centering
				\raisebox{-.8\height}{\includegraphics[width=\linewidth,height=.25\linewidth]{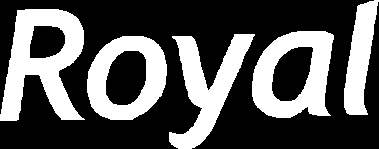}}
			\end{minipage}
		} & \raisebox{-.5\height}{Royal}& \makecell[c]{
			\raisebox{-.5\height}{\textcolor[rgb]{1,0,0}{Dyd}al}\\(no mask)\\ \raisebox{-.5\height}{Royal}\\(mask)
		} \\
		\midrule
		Ours(DAN)& \makecell[c]{
			\begin{minipage}[b]{0.2\columnwidth}
				\centering
				\raisebox{-.5\height}{\includegraphics[width=\linewidth,height=.25\linewidth]{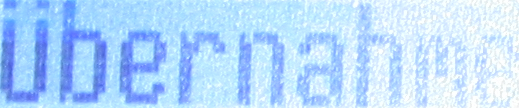}}
			\end{minipage} \\
			\begin{minipage}[b]{0.2\columnwidth}
				\centering
				\raisebox{-.8\height}{\includegraphics[width=\linewidth,height=.25\linewidth]{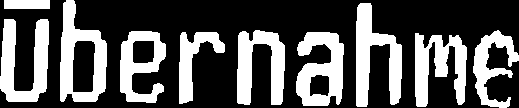}}
			\end{minipage}
		} & \raisebox{-.5\height}{ubernahme}& \makecell[c]{
			\raisebox{-.5\height}{ubernah\textcolor[rgb]{1,0,0}{ets}}\\(no mask)\\ \raisebox{-.5\height}{ubernahme}\\(mask)
		} \\
		\bottomrule
	\end{tabular}
	\caption{Comparison of the demonstration of mask effects on text recognition tasks.}
	\label{recshow}
\end{table}

Here we use ICDAR-03\cite{lucas2005icdar} and ICDAR-13 to do text recognition experiments. It can be seen from the results in Table \ref{Rec} that when we use CRNN as a text recognition model, our input has higher accuracy than the original input image, which is improved by 1.88\% on dataset ICDAR-03. Following the same experimental steps, we compare the recognition promotion effect of our mask and TSMGN's mask on the CRNN model, and we have improved accurancy by 1.47\% over them on ICDAR-13. When applying our model to DAN, the accurancy improvement is 0.35\% and 0.51\% on the ICDAR-03 and ICDAR-13 datasets, respectively. 

\begin{figure}[h]
	\centering
	\includegraphics[width=0.95\linewidth]{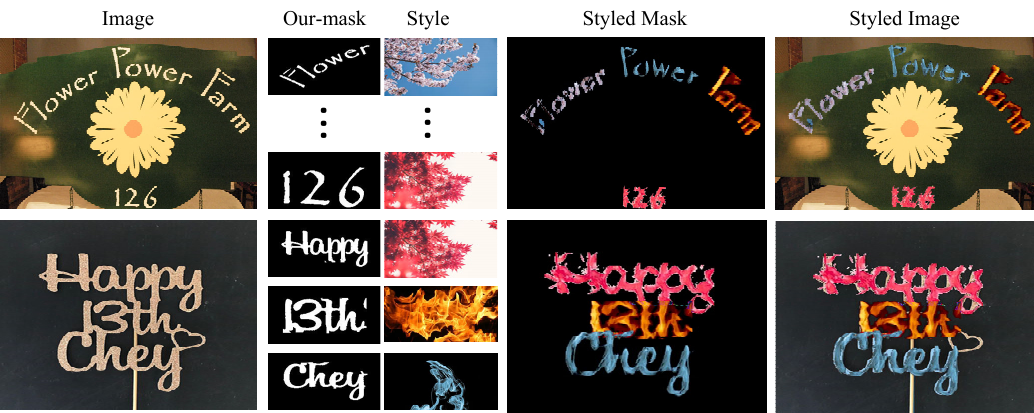}
	\caption{Examples of text style transfer with different styles by our method and SMG.}
	\label{style}
\end{figure}

\noindent \textbf{Text style transfer:}\ As a trend in mechanical design, text style transfer has marvelous value in industry and research. Like TexRNet, we also use SMG (Shape Matching GAN) \cite{yang2019controllable} as our downstream application, requiring text masks and style images as input. But unlike TexRNet, which takes the mask of the entire image as input for consistent style transfer across all instances, we take the mask of each local text instance as input. So we can have more custom designs for different text instances in an image. Our example of text style transfer is shown in Figure  \ref{style}.

\begin{table}
	\begin{tabular}{ccc}
		\toprule
		Model&ICDAR-03&ICDAR-13\\
		\midrule
		CRNN & 89.40& 86.70\\
		TSMGN(CRNN)& \textbf{91.35($\uparrow$1.95)} & 87.86($\uparrow$0.98)\\
		Ours(CRNN)& 91.28($\uparrow$1.88)& \textbf{89.15($\uparrow$2.45)}\\
		DAN& 95.00& 93.90\\
		Ours(DAN)& \textbf{95.35($\uparrow$0.35)}& \textbf{94.41($\uparrow$0.51)}\\
		\bottomrule
	\end{tabular}
	\caption{Comparison of the promotion effects of masks generated by different methods on text recognition tasks.}
	\label{Rec}
\end{table}

\section{Conclusion}
In this paper, in order to alleviate the problems of the poor quality and high cost of annotation of pixel-level the current scene text segmentation datasets and the enormous gap in data distribution of pretraining transfer, we first propose an object-centric self-supervised scene text segmentation algorithm without using pixel-level labels and pretraining on synthetic datasets.
We propose a text region query module and representation consistency constraints according to text regions inferred by the text localization model. Compared with general unsupervised segmentation algorithms, our method has a certain improvement. As for limitations, we hope to utilize less human cost in the future, such as replacing polygon text regions with bounding box text regions. At the same time, incorporating a top-down approach into a bottom-up approach is also a direction we need to proceed.

\bibliographystyle{ACM-Reference-Format}
\bibliography{sample-base}

\end{document}